\documentclass[conference]{IEEEtran}

\IEEEoverridecommandlockouts
\usepackage{amsmath,amssymb,amsfonts}
\usepackage{times}
\usepackage{cite}
\usepackage[hyphens]{url}
\usepackage{graphicx}
\usepackage{epsfig}
\usepackage{textcomp}
\usepackage{balance}
 
\usepackage[table]{xcolor}
\usepackage{tabularx}
\usepackage{booktabs}
\usepackage{multirow}
\usepackage{array}
\usepackage{pifont}
\usepackage{subcaption}
\usepackage{collcell}
\usepackage{pgfmath}
\usepackage{multirow}
\usepackage{algorithm}
\usepackage{algpseudocode}
\usepackage{ulem}

\definecolor{cbBlue}{RGB}{0,114,178}
\definecolor{cbOrange}{RGB}{230,159,0}

\newcommand{\GC}[1]{%
    \pgfmathparse{#1 > 60 ? 1 : 0}%
    \ifnum\pgfmathresult=1
        \cellcolor{cbBlue!#1!white}\textcolor{white}{#1}%
    \else
        \cellcolor{cbBlue!#1!white}\textcolor{black}{#1}%
    \fi
}

\newcommand{\RC}[1]{\cellcolor{cbOrange!#1!white}\textcolor{black}{#1}}

\begin{document}

\title{Detecting LLM-Assisted Vietnamese Writing \\ via Keystrokes under Behavioral Manipulation}



\author{
\IEEEauthorblockN{Thanh Dong*}
\IEEEauthorblockA{
\textit{Bucknell University}\\
Lewisburg, PA, USA\\
tpd010@bucknell.edu
}
\and
\IEEEauthorblockN{An Ngo*}
\IEEEauthorblockA{
\textit{Bucknell University}\\
Lewisburg, PA, USA\\
axn001@bucknell.edu
}
\and
\IEEEauthorblockN{Minh Dau}
\IEEEauthorblockA{
\textit{Bucknell University}\\
Lewisburg, PA, USA\\
mtcd001@bucknell.edu
}
\and
\IEEEauthorblockN{Rajesh Kumar}
\IEEEauthorblockA{
\textit{Bucknell University}\\
Lewisburg, PA, USA\\
rajesh.kumar@bucknell.edu
}
\thanks{*Thanh Dong and An Ngo contributed equally to this work.}
}
\maketitle
\begin{abstract}
 We study the robustness of keystroke dynamics for detecting large language model (LLM)-assisted writing. We introduce a Vietnamese keystroke dataset capturing realistic writing modes, including bona fide composition, transcription, and paraphrasing. We also define a behaviorally grounded threat model in which users deliberately alter typing patterns. To implement the threat model, we create behaviorally manipulated variants of the data designed to evade keystroke-based detection. We evaluate four keystroke modeling approaches: temporal and rhythmic representations, and sequential representations modeled with a one-dimensional convolutional neural network (1D-CNN) and TypeNet, under user-independent and context-independent settings. The results show that sequential models outperform feature-based approaches in most cases and that keystroke signals encode discriminative information about the writing process. However, detection is not uniformly robust: transcription is reliably identified, while paraphrasing and adversarially manipulated samples are frequently misclassified as bona fide when not explicitly modeled. To address this, we incorporate adversarial training using behaviorally manipulated data, which substantially improves separability and robustness. These results suggest that keystroke-based detection depends critically on exposure to diverse writing behaviors, and that strong performance under limited conditions does not generalize to realistic or adversarial settings without targeted modeling.
\end{abstract}

\begin{IEEEkeywords}

Keystroke Dynamics, LLM-Assisted Writing Detection, 
Adversarial Robustness, Human vs. LLM

\end{IEEEkeywords}
 
\section{Introduction}
\label{sec:introduction}
Recent advances in large language models (LLMs) such as ChatGPT \cite{ChatGPT}, Claude \cite{Claude}, Gemini \cite{geminiteam2025}, and LLaMA \cite{grattafiori2024llama} have made it increasingly easy to generate high-quality text. This capability allows students to generate, paraphrase, or transcribe responses with minimal effort, raising significant concerns about academic integrity, particularly in online and take-home assessment settings \cite{kundu2024, Roh2025, crossley2024, mehta2025}. Text-based approaches have also explored stylometric features for detecting LLM-assisted writing, demonstrating that lexical and grammatical characteristics can provide useful discriminative signals \cite{kumar2027stylometric}. However, such approaches analyze the final written output rather than the behavioral process through which it is produced. Traditional plagiarism detection systems based on textual similarity can also become less effective when LLM-generated content is substantially paraphrased or rewritten \cite{kumarage2024, alzahrani2012}.

Prior studies further indicate that human evaluators struggle to reliably distinguish between human-written and artificial intelligence (AI)-generated text under such conditions, with performance approaching random guessing \cite{CoinTossHuman, mehta2025}.

Rather than replacing conventional plagiarism detection systems, we view keystroke-based analysis as a complementary \underline{layer} of academic integrity assessment. In a practical deployment, submissions would first be screened using similarity-based plagiarism detectors, which remain effective for identifying copied material from known or indexed sources. Submissions that pass this initial screening but remain suspicious because they may have been generated or heavily paraphrased using LLMs can then be analyzed using keystroke dynamics, which examine the writing process rather than the final text.

These limitations motivate a shift from analyzing the final written output to examining the process through which the text is produced \cite{kundu2024, Roh2025, crossley2024, mehta2025}. Keystroke dynamics capture fine-grained behavioral signals during text production, including typing speed, pauses, and revision patterns \cite{kundu2024, Roh2025, crossley2024, mehta2025}. Such signals reflect underlying cognitive and motor processes involved in writing, including planning, translation, and editing \cite{condrey2026detectingcognitive}. Unlike textual features, which can be manipulated through paraphrasing or rewriting, keystroke signals arise from real-time interaction between cognition and motor execution, making them inherently difficult to replicate without altering the writing process. Altering the writing process and corresponding keystrokes is non-trivial \cite{HassanKhanKeystrokeImitation}. Prior work has demonstrated systematic differences between bona fide and AI-assisted writing in these behavioral patterns, suggesting that keystroke dynamics provide a reliable signal for detecting LLM-assisted content \cite{kundu2024, crossley2024, mehta2025, Roh2025}.

Despite this promise, existing work remains limited in several important respects \cite{mehta2025}. First, most approaches consider simplified problem settings, typically distinguishing only between bona fide and assisted writing \cite{kundu2024, crossley2024, mehta2025, Roh2025}. In practice, LLM assistance can take different forms: a user may directly transcribe an LLM-generated response or paraphrase it while composing the final answer. These interaction modes differ in cognitive demand and consequently produce distinct behavioral signatures in keystroke dynamics \cite{Roh2025, mehta2025}. Importantly, paraphrasing may reduce textual evidence of LLM assistance while preserving behavioral differences in the writing process.

Second, the problem becomes adversarial at the behavioral level when detection relies on keystroke signals. In this setting, users may attempt to evade detection not by modifying the textual output \cite{Jemama2025}, but by deliberately altering \emph{how} they type. This includes introducing artificial pauses, revisions, or variations in typing speed to mimic natural writing behavior. While prior work \cite{Roh2025} has explored adversarial settings using synthetic statistical perturbations, such approaches do not fully capture behaviorally grounded evasion, where users consciously adapt their typing process during composition. This distinction separates task-level variation (transcription vs.\ paraphrasing) from strategic behavioral manipulation aimed at bypassing keystroke-based detectors.

Third, the applicability of keystroke-based detection across languages remains poorly understood \cite{Roh2025}. Existing studies focus primarily on English \cite{kundu2024, crossley2024, agarwal2022contract} and, more recently, Korean \cite{Roh2025}. It remains unclear whether these methods generalize to languages with different typing characteristics, input methods, and orthographic structures. 

Vietnamese provides a particularly informative testbed because its TELEX input method composes diacritics through multi-keystroke sequences, resulting in typing dynamics that differ substantially from English and Korean \cite{telex_input,vietnamese_input_methods, vietnamese_keyboard}. This allows us to evaluate whether existing keystroke-based detection methods generalize across fundamentally different input mechanisms. Thus, Vietnamese provides a distinct setting for evaluating keystroke-based detection, while generalization to other languages remains to be investigated.
  
To bridge these gaps, we develop and evaluate a behavioral detection framework that complements conventional text-based plagiarism detection for Vietnamese LLM-assisted writing. In particular, this work makes the following contributions: 

\begin{itemize}
    \item A Vietnamese keystroke dynamics dataset for analyzing AI-assisted writing across multiple realistic scenarios, publicly available for research on plagiarism detection and keystroke-based authentication. \footnote{https://github.com/dongphuthanh/VietnameseKeystrokes}
    
    \item A comparative evaluation of temporal, rhythmic, and deep sequential modeling approaches, including a one-dimensional convolutional neural network (1D-CNN) and TypeNet.
    
    \item An analysis of detection performance across diverse writing conditions, including synthetically generated behavioral manipulations designed to simulate adversarial evasion strategies.
    
    \item An evaluation of adversarial training for improving robustness against these synthetically generated behavioral manipulations.
\end{itemize}

The remainder is organized as follows: Section~\ref{sec:related_work} reviews related work, Section~\ref{sec:methodology} presents the approach, Section~\ref{sec:results_discussion} reports results, and Section~\ref{sec:conclusion_and_futurework} concludes.  

\begin{table*}[t]
\caption{
Comparison of keystroke-based approaches across language (EN: English, KR: Korean, VI: Vietnamese), representation (rhythmic, temporal, deep), classifiers (RF: random forest, ML: classical machine learning), threat assumptions, and writing modes (B: bona fide, T: transcribe LLM, P: paraphrase LLM, FT/FP: adversarial variants).
}

\vspace{0.1in}
\centering
\small
\setlength{\tabcolsep}{5pt}
\renewcommand{\arraystretch}{1.05}

\begin{tabularx}{\textwidth}{
l
c
c
X
c
c
c
}
\toprule
\textbf{Work} & \textbf{Lang} & \textbf{Cognition} & \textbf{Modeling} & \textbf{Classifier} & \textbf{Threat} & \textbf{Modes} \\
\midrule

Crossley et al.~\cite{crossley2024} 
& EN 
& Implicit 
& Rhythmic
& RF 
& None 
& \{B, T\} \\

Kundu et al.~\cite{kundu2024} 
& EN 
& None 
& Deep 
& TypeNet 
& None 
& \{B, T\} \\

Roh et al.~\cite{Roh2025} 
& KR 
& Explicit
& Rhythmic, Temporal 
& ML 
& None 
& \{B, T, P\} \\

Mehta et al.~\cite{mehta2025} 
& EN 
& Implicit 
& Temporal, Deep 
& TypeNet, ML 
& Statistical
& \{B, T, P\} \\

\textbf{Ours} 
& {VI} 
& {Explicit} 
& {Rhythmic, Temporal, Deep}
& {ML, 1D-CNN, TypeNet}
& {Behavioral} 
& \{B, T, P, FT, FP\} \\

\bottomrule
\end{tabularx}
\label{tab:related_work_comparison}
\end{table*}
 

\section{Related Work}
\label{sec:related_work}
Keystroke dynamics has been widely studied as a behavioral signal for user authentication, identification, and fraud detection \cite{joyce1990, leggett1988, fabianmonrose2000, kumar2016, kumar2018, KeystrokeVideo, KeystrokeSound, kuruvilla2024, kim2026you}. These works show that typing patterns encode stable temporal characteristics that can be used to distinguish users in different contexts. Beyond identity modeling, keystroke behavior has also been explored as a proxy for cognitive processes and as complementary information for writing assessment \cite{schaller2026}. Banerjee et al.~\cite{banerjee2014keystroke} study keystroke patterns as an analog to prosody in speech to capture the real-time writing process for deception detection. They collect a dataset of truthful and deceptive texts, extract features related to editing behavior and temporal dynamics such as pauses and typing speed, and incorporate them into a classification framework. Their results show that keystroke-based features improve deception detection performance, indicating that differences in cognitive load between truthful and deceptive writing are reflected in measurable typing patterns.

Agarwal et al.~\cite{agarwal2022contract} design a keystroke-based system to distinguish genuine students from impostors in remote assessments by modeling typing behavior as a biometric signal. The approach uses duration and latency features and evaluates detection performance under strict constraints such as zero false accusations. Their results show that keystroke dynamics can effectively identify contract cheating, supporting the use of behavioral signals for verifying authorship in academic settings. More recently, Crossley et al.~\cite{crossley2024} used keystroke logs to distinguish authentic writing from transcribed text. Their model leverages rhythmic features such as pauses, revisions, and writing bursts to capture differences in writing processes and achieves high classification accuracy. Similarly, Deane et al.~\cite{TypeNetNotInterpretableETS} used keystroke dynamics to distinguish original composition from nonoriginal text production, further demonstrating that differences in the writing process can provide signals beyond the final text. Zhang et al.~\cite{ZHANG2026} further examined natural writing and copy typing using keystroke logs and deep learning models, showing distinct behavioral patterns between the two writing processes. However, these studies focus on transcription, copy typing, or nonoriginal text production rather than direct interaction with LLMs during composition.

Recent work explicitly studies LLM-assisted writing rather than treating transcription or authorship mismatch as a proxy \cite{kundu2024, Roh2025, mehta2025, Asher2026}. These studies share a common goal of distinguishing bona fide writing from interactions with generative models, but differ in how they model LLM usage and behavioral variability. Early work \cite{kundu2024} formulates the problem as binary classification between bona fide and ChatGPT-assisted writing using deep sequential models such as TypeNet \cite{typenet}. While this demonstrates that keystroke dynamics capture differences in writing processes, the formulation assumes a single mode of LLM assistance and does not distinguish between behaviors such as transcription and paraphrasing.

Subsequent work from Roh et al.~\cite{Roh2025} separates ChatGPT interaction into transcription and paraphrasing modes and incorporates cognitive variation using Bloom’s Taxonomy. Their results show that paraphrasing produces more variable and complex keystroke patterns than transcription, highlighting the role of cognitive load in shaping behavior. Mehta et al.~\cite{mehta2025} further extend this line of work by scaling the dataset, introducing explicit paraphrasing conditions, and benchmarking keystroke-based models against text-based detectors and human evaluators. They also introduce a deception-oriented threat model and evaluate robustness under adversarial settings. In their formulation, an adversary generates forged keystroke sequences by transferring timing statistics from prior user data to new text, simulating attacks such as user-specific and pooled timing-based mimicry. Their results show that such attacks significantly degrade model performance and that adversarial training improves robustness to these perturbations.

However, the proposed threat model operates primarily at the level of statistical timing synthesis and does not fully capture how users may adapt their typing behavior during real-world typing. In practice, adversaries may introduce higher-level behavioral strategies, such as modifying pause patterns, revision behavior, or writing rhythm, which are not explicitly modeled in their framework. As a result, while the study provides an important first step toward evaluating deception, it does not cover the full space of behaviorally grounded adversarial strategies.

As summarized in Table~\ref{tab:related_work_comparison}, prior approaches are limited in modeling diversity, adversarial realism, and linguistic diversity, with existing studies focusing primarily on English \cite{kundu2024, mehta2025} and Korean \cite{Roh2025}. The lack of language diversity raises questions about whether keystroke-based signals generalize across typologically distinct languages, such as Vietnamese, which exhibits different lexical segmentation and input dynamics. This work addresses these gaps by introducing a Vietnamese keystroke dataset with multiple writing modes, including transcription and paraphrasing, as well as a behaviorally grounded threat model for realistic adversarial behavior. In addition, we incorporate adversarial training and evaluate across temporal, rhythmic, and sequence-based representations.


\section{Methodology}
\label{sec:methodology}
\subsection{Task definition}
\label{task_description}
Detecting varying levels of LLM intervention in the writing process can be formulated as a multi-class classification problem. A user produces a text response \(t\) through a sequence of keystroke events \(k\), where the keystroke sequence captures the underlying writing behavior. The objective is to infer the type of writing process that generated the response, distinguishing between genuine composition and different forms of assisted or deceptive writing.

Each sample is represented as a pair \((t, k)\), where \(t\) denotes the final text and 
\( 
k = \{(k_i, t_i^{\downarrow}, t_i^{\uparrow})\}_{i=1}^{N}
\)
denotes the corresponding keystroke sequence. Here, \(k_i\) represents the key identity at event \(i\), \(t_i^{\downarrow}\) denotes the timestamp at which the key is pressed (key-down event), and \(t_i^{\uparrow}\) denotes the timestamp at which the key is released (key-up event). These events encode fine-grained temporal dynamics of the writing process, including key hold times and inter-key latencies.

Each sample is associated with a label \(y \in \mathcal{Y}\), where \( 
\mathcal{Y} = \{B, T, P, FT, FP\}.\) The label \(B\) denotes bona fide writing, where the text is independently composed without external assistance. The remaining classes correspond to different forms of deviation from genuine writing. The labels \(T\) and \(P\) denote text-level interventions in which the user relies on LLM-generated content. \(T\) is for direct transcription (T) and \(P\) is for paraphrasing (P) without any intentional alteration to the typing process. 

The labels \(FT\) and \(FP\) denote behavioral-level interventions in which, while transcribing or paraphrasing, the user intentionally modifies keystroke behavior to mimic a bona fide writing process or to fool the detectors. These modifications include changes in typing speed, pause structure, and revision patterns, with the goal of deceiving the detectors by producing keystroke traces that resemble genuine composition.

We learn a classifier \( f_\theta: \mathcal{X} \rightarrow \mathcal{Y} \), where \(\mathcal{X}\) denotes the space of keystroke sequences. The model operates on representations derived solely from \(k\).

\subsection{Detection framework overview}

The proposed framework assumes that conventional text-based plagiarism screening has already been performed. Consequently, our detector is designed to analyze submissions that exhibit little textual similarity to existing sources but may nevertheless involve substantial LLM assistance. In this deployment scenario, keystroke dynamics provides a complementary behavioral analysis after the initial text-based screening stage.

In a practical deployment, keystrokes could be captured within a web-based assessment interface, as in our data-collection system, without requiring background software. Collection would be limited to the assessment session and would require appropriate disclosure and privacy safeguards.

The detection framework consists of six stages: data collection, preprocessing, adversarial generation, behavioral modeling and representation, training, and evaluation. Each stage is described in detail below. 

\subsubsection{Data collection}
Following Institutional Review Board (IRB) approval from the university, a total of $45$ Vietnamese participants were recruited, including $19$ university students and $26$ high school students, with $11$ females and $34$ males (mean age $17.66$ years). The study consisted of two phases separated by at least two weeks; $30$ participants completed both phases. All participants provided informed consent prior to participation, with appropriate procedures for minor participants.

Participants completed the study on personal devices to preserve typing familiarity. Data were collected in real-world environments, including classrooms, cafés, and residences, introducing natural variations in typing conditions. One author was present for all sessions to resolve technical issues and ensure adherence to the protocol.

The protocol included three writing modes following prior work \cite{Roh2025}: bona fide composition (B), transcription of LLM-generated responses (T), and paraphrasing of LLM-generated content (P). Prompts were carefully crafted to include questions of varying cognitive demand, and responses were constrained to $90$--$150$ words \cite{Roh2025}. This range provides sufficiently long keystroke sequences for behavioral analysis while limiting participant fatigue and maintaining comparable response lengths across conditions. For T and P, participants first generated LLM responses that satisfied the length constraint and then performed the task.

Keystroke events were recorded using a web-based interface, capturing key identity and key-down/key-up timestamps. For T and P, a reference field displays the LLM response, while a separate input field records user-generated text, ensuring complete keystroke traces. Copy-paste was disabled because the study targets exam settings and the proposed method requires observing the writing process. Handling pasted and subsequently edited text is left for future work, as it introduces varying levels of user modification that are not captured in the current protocol. 

All responses were subjected to automated validation. Lexical validity required at least $85\%$ of tokens to appear in a Vietnamese lexicon. Transcription responses were required to achieve $\geq 95\%$ similarity to the source text. Paraphrased responses were required to exhibit $\geq 15\%$ lexical divergence. In addition, responses were evaluated using the Gemini $2.5$ Flash application programming interface (API) \cite{deepmind2024gemini25}, which scores relevance, coherence, depth, and grammatical quality on a $0$--$3$ scale per dimension; a minimum total score of $7$ (out of $12$) was required for inclusion. Participants were informed of these criteria prior to the study to replicate a real exam environment, and no responses failed the validation thresholds. These criteria served as quality-control measures to remove invalid responses, ensure close adherence in transcription, and require sufficient modification in paraphrasing.

\subsubsection{Data cleaning and preprocessing}
Raw keystroke logs were normalized to ensure consistency across input systems and to remove noise prior to representation learning. Vietnamese text is typically entered using TELEX, a diacritic input method. Since the dataset was collected from heterogeneous devices, including Windows laptops and MacBooks, discrepancies arise in recorded keystroke sequences. On macOS, TELEX is natively supported and produces keystroke logs that closely reflect user-intended inputs. In contrast, most Windows users rely on Unikey, which introduces intermediate virtual keystrokes when composing diacritics.

For example, typing ``chào'' requires the sequence \textit{c--h--a--f--o}, where \textit{f} transforms \textit{a} into à. Crucially, the same lexical output can also be produced via the alternative ordering \textit{c--h--a--o--f}, reflecting a non-canonical keystroke permutation characteristic of this input method, with admissible variations in ordering across user input strategies. Under Unikey, additional transient keystrokes are generated between \textit{a} and \textit{f}, with patterns varying across software versions. These artifacts do not correspond to user-level actions and distort temporal and sequential features.

We identify recurring Unikey-specific patterns and remove virtual keystrokes through an automated filtering procedure, yielding sequences that retain only user-intended inputs.

Following normalization, all keystroke sequences undergo a unified two-stage noise filtering procedure applied across all modeling approaches. Let $K(x)=\{(k_i, t_i^{\downarrow}, t_i^{\uparrow})\}_{i=1}^{N}$, with derived features including key hold time $h_i = t_i^{\uparrow}-t_i^{\downarrow}$, inter-key interval $\Delta_i = t_{i+1}^{\downarrow}-t_i^{\downarrow}$, and release-to-keydown time (RKDT).

In the first stage, timing values are constrained to a physiologically plausible range. Any $h_i$, $\Delta_i$, or related timing feature outside $[50, 5000]$ ms is removed, eliminating artifacts such as accidental key holds or system-level logging delays.

In the second stage, outliers are removed using an interquartile range (IQR) filter applied to each timing feature distribution. Specifically, values outside $[Q1 - 2 \cdot IQR,\, Q3 + 2 \cdot IQR]$ are discarded to suppress anomalous pauses while preserving natural variability.

\subsubsection{Threat model and adversarial sample generation}

We consider the manipulation of keystroke dynamics during LLM-assisted writing. In this setting, a user produces an LLM-assisted response by typing, while deliberately altering their typing behavior to evade detection.

Let $(t,k,y)$ denote a sample as defined in Section~\ref{task_description}, where $k = \{(k_i, t_i^{\downarrow}, t_i^{\uparrow})\}_{i=1}^{N}$ is the keystroke sequence and $y \in \mathcal{Y} = \{B, T, P\}$ is the writing mode. Keystroke events are represented as paired keydown/keyup events, from which key hold times $h_i = t_i^{\uparrow} - t_i^{\downarrow}$ and inter-key intervals $\Delta_i = t_{i+1}^{\downarrow} - t_i^{\downarrow}$ are derived. The detector $f_\theta : \mathcal{X} \rightarrow \mathcal{Y}$ operates only on $k$.

For samples with $y \in \{T, P\}$, we assume that users may deliberately modify their typing behavior during transcription and paraphrasing to evade detection. Such manipulation produces a keystroke sequence $\hat{k}$ during typing that deviates from natural transcription and paraphrasing patterns and more closely resembles bona fide writing. The objective is to induce misclassification as bona fide, i.e., $f_\theta(\hat{k}) = B$.

We assume a zero-knowledge black-box setting in which the adversary has no access to model parameters or features, but is aware that transcription and paraphrasing produce regular, low-variance timing patterns with limited revisions. These differences define the key behavioral gap between LLM-assisted and bona fide writing: transcription and paraphrasing exhibit reduced temporal variability, minimal revisions, and weaker pause structure \cite{crossley2024, Roh2025}. The adversarial transformations are designed to explicitly target these deficiencies.

The manipulation operates on the temporal structure of $k$. The adversary perturbs $\{h_i, \Delta_i\}$ during typing to alter higher-order statistics, including variance, burst structure, pause distribution, and revision frequency. 

In practice, we approximate this behavior by constructing adversarial samples through a transformation $\mathcal{A}: \mathcal{X} \rightarrow \mathcal{X}$ applied to recorded keystroke sequences. For each $(t,k,y)$ with $y \in \{T,P\}$, we generate
\( 
\hat{k} = \mathcal{A}(k), \quad k \in \mathcal{X}, \; \hat{k} \in \mathcal{X}
\)
and assign new labels $FT$ and $FP$ to the behavioral-level interventions defined in Section~\ref{task_description}. The transformation $\mathcal{A}$ operates on $\{h_i,\Delta_i\}$ and consists of three components: temporal variability (velocity scaling), revision behavior (revision injection), and cognitive planning signals (boundary-aware pausing). To induce \underline{temporal variability}, typing speed is modulated by scaling $\Delta_i$ with a factor $v \sim U(0.75,1.10)$, resampled at word boundaries with probability $0.15$, increasing variance and disrupting the near-constant rhythm observed in transcription, which lacks natural motor variability.

The \underline{revision behavior} is manipulated by deleting and retyping segments, which are introduced with probability $0.10$ at word or sentence boundaries, ensuring at least two revision episodes per sequence. Each revision deletes $2$--$5$ words via backspace events, preceded by a hesitation delay sampled from $U(600,1400)$ ms and followed by a re-planning delay sampled from $U(900,2600)$ ms. Subsequent backspace events occur with short inter-key intervals ($60$--$110$ ms), modeling rapid motor execution following error realization. Retyped text is produced with reduced latency ($\Delta_i \leftarrow 0.7\Delta_i$, $h_i \leftarrow 0.85 h_i$), reflecting motor fluency during repetition.

To accomplish boundary-aware pausing, additional delays were inserted into $\Delta_i$ at linguistic boundaries with probabilities $0.20$ (word), $0.45$ (clause), $0.70$ (sentence), and $0.85$ (line). Pause durations were sampled from Weibull distributions (mean $\approx 750$--$1750$ ms, shape $k \in [2.1,2.4]$), with a long-tail mixture (probability $0.4$--$0.6$, multiplier $2.5$--$3.0$) to model occasional extended planning pauses. Pauses were not injected when $\Delta_i > 1500$ ms.

\subsubsection{Behavioral representation and modeling}

Each preprocessed sequence $\tilde{k}$ is mapped to
\( z = \{\phi_{\text{temp}}(\tilde{k}), \; \phi_{\text{rhythm}}(\tilde{k}), \; \phi_{\text{seq}}(\tilde{k})\},
\)
capturing complementary behavioral signals at motor, structural, and sequential levels.

The temporal representation $\phi_{\text{temp}}(\tilde{k})$ captures low-level typing dynamics through key hold time (KHT), key interval time (KIT), and release-to-keydown time (RKDT), defined as $h_i = t_i^{\uparrow} - t_i^{\downarrow}$, $\Delta_i = t_{i+1}^{\downarrow} - t_i^{\downarrow}$, and $r_i = t_{i+1}^{\downarrow} - t_i^{\uparrow}$, respectively. Features are computed over the top-$50$ most frequent digraphs \cite{sim2007digraphs} to reduce sparsity and focus on stable transitions, and summarized using
\( 
\{\mu, \sigma, \sigma/\mu, \text{range}, \max, \text{median}\},
\)
yielding a fixed-length representation (with dimensionality determined by the number of digraph-feature combinations) suitable for tabular models.

The rhythmic representation $\phi_{\text{rhythm}}(\tilde{k})$ consists of $107$ features derived from pause, burst, and revision statistics \cite{Roh2025, crossley2024}. These include binned pause durations; inter-word and inter-sentence pauses; pauses preceding deletions; P-bursts (production bursts), R-bursts (revision bursts), and deletion bursts; and distributional descriptors such as moments, quantiles, counts, and entropy. These features are complementary to temporal statistics and capture long-range structural patterns not encoded in local timing signals.

The sequential representation $\phi_{\text{seq}}(\tilde{k})$ models keystroke dynamics as an ordered signal with events encoded as $x_i = (\Delta_i, c_i)$, where $c_i$ denotes the key category. To avoid leaking textual content and prevent models from exploiting lexical information, keys are mapped into five classes: characters, digits, space, backspace, and special keys, while preserving structural cues such as word boundaries and deletions. Sequences are segmented using sliding windows of length $L=100$ (key-down events) with a stride of $50$, selected to balance temporal resolution and contextual coverage.

A 1D-CNN is used to model local temporal dependencies. To handle heterogeneous keystroke data, continuous timing and discrete keycodes are first processed by a multilayer perceptron (MLP) and an embedding layer, respectively. Their outputs are concatenated and fed into a secondary MLP fusion layer to create a unified, high-dimensional representation before entering the convolutional stack. This sequence is then fed into three stacked convolutional layers with a kernel size of $3$, capturing trigram-level keystroke patterns that reflect short-range motor dependencies \cite{1DCNN}. The convolutional stack is followed by global max pooling, which provides invariance to sequence length and captures salient timing anomalies, and by a two-layer fully connected classification head that maps learned representations to class probabilities. We additionally employ TypeNet \cite{typenet}, an established two-layer long short-term memory (LSTM) network trained with cross-entropy, as a recurrent baseline; while the 1D-CNN captures local temporal dependencies, the LSTM models longer-range sequential structure across keystrokes. The sequential representation for TypeNet is modeled as $x_i=(t_{i+1}^{\downarrow} - t_i^{\downarrow},\ t_i^{\uparrow} - t_i^{\downarrow},\ t_{i+1}^{\downarrow} - t_i^{\uparrow},\ t_{i+1}^{\uparrow} - t_{i}^{\uparrow},\ c_i) $, following the original TypeNet design \cite{typenet}.

Given $z$, classifiers $f \in \mathcal{F}$ are trained under each evaluation setting. Temporal and rhythmic representations are modeled using eXtreme Gradient Boosting (XGBoost), while $\phi_{\text{seq}}$ is modeled using 1D-CNN and recurrent architectures. For sequence models, predictions are computed at the window level and aggregated via mean pooling across windows, followed by class selection based on maximum probability to obtain the final prediction $\hat{y}$.


\begin{figure*}[t]
    \centering
    \renewcommand{\arraystretch}{0.83}
    \setlength{\tabcolsep}{1pt} 
    
    \resizebox{\textwidth}{!}{
    
    \begin{tabular}{c | c | c | c}
    
    \multicolumn{4}{c}{\textbf{Rhythmic - User Independent}} \\ [1ex]
    $\begin{array}{c|ccccc}
         & \multicolumn{1}{c}{\text{B}} & \multicolumn{1}{c}{\text{P}} & \multicolumn{1}{c}{\text{T}} & \multicolumn{1}{c}{\text{FP}} & \multicolumn{1}{c}{\text{FT}} \\ \cline{2-6}
        \text{B} & \GC{88.4} & \GC{0.0} & \GC{11.6} & \GC{0.0} & \GC{0.0} \\
        \text{P} & \RC{74.2} & \RC{0.0} & \RC{25.8} & \RC{0.0} & \RC{0.0} \\
        \text{T} & \GC{13.3} & \GC{0.0} & \GC{86.7} & \GC{0.0} & \GC{0.0} \\
        \text{FP} & \RC{85.9} & \RC{0.0} & \RC{14.1} & \RC{0.0} & \RC{0.0} \\
        \text{FT} & \RC{36.3} & \RC{0.0} & \RC{63.7} & \RC{0.0} & \RC{0.0}
    \end{array}$
    &
    $\begin{array}{ccccc}
        \multicolumn{1}{c}{\text{B}} & \multicolumn{1}{c}{\text{P}} & \multicolumn{1}{c}{\text{T}} & \multicolumn{1}{c}{\text{FP}} & \multicolumn{1}{c}{\text{FT}} \\ \hline
        \GC{64.4} & \GC{30.7} & \GC{4.9} & \GC{0.0} & \GC{0.0} \\
        \GC{40.4} & \GC{43.6} & \GC{16.0} & \GC{0.0} & \GC{0.0} \\
        \GC{7.1} & \GC{15.6} & \GC{77.3} & \GC{0.0} & \GC{0.0} \\
        \RC{41.5} & \RC{54.8} & \RC{3.7} & \RC{0.0} & \RC{0.0} \\
        \RC{26.7} & \RC{43.0} & \RC{30.4} & \RC{0.0} & \RC{0.0}
    \end{array}$
    &
    $\begin{array}{ccccc}
        \multicolumn{1}{c}{\text{B}} & \multicolumn{1}{c}{\text{P}} & \multicolumn{1}{c}{\text{T}} & \multicolumn{1}{c}{\text{FP}} & \multicolumn{1}{c}{\text{FT}} \\ \hline
        \GC{57.3} & \GC{36.9} & \GC{5.8} & \GC{0.0} & \GC{0.0} \\
        \GC{32.0} & \GC{50.2} & \GC{17.8} & \GC{0.0} & \GC{0.0} \\
        \GC{7.1} & \GC{20.0} & \GC{72.9} & \GC{0.0} & \GC{0.0} \\
        \RC{0.0} & \RC{0.0} & \RC{0.0} & \RC{0.0} & \RC{100.0} \\
        \GC{0.0} & \GC{0.0} & \GC{0.0} & \GC{0.0} & \GC{100.0}
    \end{array}$
    &
    $\begin{array}{ccccc}
        \multicolumn{1}{c}{\text{B}} & \multicolumn{1}{c}{\text{P}} & \multicolumn{1}{c}{\text{T}} & \multicolumn{1}{c}{\text{FP}} & \multicolumn{1}{c}{\text{FT}} \\ \hline
        \GC{60.0} & \GC{33.3} & \GC{6.7} & \GC{0.0} & \GC{0.0} \\
        \GC{34.7} & \GC{48.9} & \GC{16.4} & \GC{0.0} & \GC{0.0} \\
        \GC{6.2} & \GC{15.6} & \GC{78.2} & \GC{0.0} & \GC{0.0} \\
        \GC{0.0} & \GC{0.0} & \GC{0.0} & \GC{74.8} & \GC{25.2} \\
        \GC{0.0} & \GC{0.0} & \GC{0.0} & \GC{23.0} & \GC{77.0}
    \end{array}$
\\[1.5ex]
    M2 & M3 & M4 & M5 \\[1.5ex]
    
\multicolumn{4}{c}{\textbf{Temporal - User Independent}} \\ [1ex]
$\begin{array}{c|ccccc}
 & \multicolumn{1}{c}{\text{B}} & \multicolumn{1}{c}{\text{P}} & \multicolumn{1}{c}{\text{T}} & \multicolumn{1}{c}{\text{FP}} & \multicolumn{1}{c}{\text{FT}} \\ \cline{2-6}
\text{B}    & \GC{64.0} & \GC{0.0}  & \GC{36.0} & \GC{0.0} & \GC{0.0} \\
\text{P}    & \RC{53.3} & \RC{0.0}  & \RC{46.7} & \RC{0.0} & \RC{0.0} \\
\text{T}    & \GC{30.7} & \GC{0.0}  & \GC{69.3} & \GC{0.0} & \GC{0.0} \\
\text{FP} & \RC{77.3} & \RC{0.0}  & \RC{22.7} & \RC{0.0} & \RC{0.0} \\
\text{FT} & \RC{38.7} & \RC{0.0}  & \RC{61.3} & \RC{0.0} & \RC{0.0}
\end{array}$
&
$\begin{array}{ccccc}
\multicolumn{1}{c}{\text{B}} & \multicolumn{1}{c}{\text{P}} & \multicolumn{1}{c}{\text{T}} & \multicolumn{1}{c}{\text{FP}} & \multicolumn{1}{c}{\text{FT}} \\ \hline
\GC{46.7} & \GC{22.7} & \GC{30.7} & \GC{0.0} & \GC{0.0} \\
\GC{29.3} & \GC{42.7} & \GC{28.0} & \GC{0.0} & \GC{0.0} \\
\GC{14.7} & \GC{18.7} & \GC{66.7} & \GC{0.0} & \GC{0.0} \\
\RC{12.0} & \RC{74.7} & \RC{13.3} & \RC{0.0} & \RC{0.0} \\
\RC{14.7} & \RC{36.0} & \RC{49.3} & \RC{0.0} & \RC{0.0}
\end{array}$
&
$\begin{array}{ccccc}
\multicolumn{1}{c}{\text{B}} & \multicolumn{1}{c}{\text{P}} & \multicolumn{1}{c}{\text{T}} & \multicolumn{1}{c}{\text{FP}} & \multicolumn{1}{c}{\text{FT}} \\ \hline
\GC{54.7} & \GC{21.3} & \GC{24.0} & \GC{0.0} & \GC{0.0} \\
\GC{26.7} & \GC{49.3} & \GC{24.0} & \GC{0.0} & \GC{0.0} \\
\GC{18.7} & \GC{24.0} & \GC{57.3} & \GC{0.0} & \GC{0.0} \\
\RC{0.0}  & \RC{0.0}  & \RC{1.3}  & \RC{0.0} & \RC{98.7} \\
\GC{0.0}  & \GC{0.0}  & \GC{1.3}  & \GC{0.0} & \GC{98.7}
\end{array}$
&
$\begin{array}{ccccc}
\multicolumn{1}{c}{\text{B}} & \multicolumn{1}{c}{\text{P}} & \multicolumn{1}{c}{\text{T}} & \multicolumn{1}{c}{\text{FP}} & \multicolumn{1}{c}{\text{FT}} \\ \hline
\GC{54.7} & \GC{24.0} & \GC{21.3} & \GC{0.0} & \GC{0.0} \\
\GC{28.0} & \GC{37.3} & \GC{34.7} & \GC{0.0} & \GC{0.0} \\
\GC{25.3} & \GC{18.7} & \GC{56.0} & \GC{0.0} & \GC{0.0} \\
\GC{0.0}  & \GC{0.0}  & \GC{1.3}  & \GC{69.3} & \GC{29.3} \\
\GC{0.0}  & \GC{0.0}  & \GC{1.3}  & \GC{21.3} & \GC{77.3}
\end{array}$
\\[1.5ex]
M2 & M3 & M4 & M5 \\[1.5ex]
    \multicolumn{4}{c}{\textbf{1D-CNN - User Independent}} \\ [1ex]
    $\begin{array}{c|ccccc}
         & \multicolumn{1}{c}{\text{B}} & \multicolumn{1}{c}{\text{P}} & \multicolumn{1}{c}{\text{T}} & \multicolumn{1}{c}{\text{FP}} & \multicolumn{1}{c}{\text{FT}} \\ \cline{2-6}
        \text{B} & \GC{97.3} & \GC{0.0} & \GC{2.7} & \GC{0.0} & \GC{0.0} \\ 
        \text{P} & \RC{73.3} & \RC{0.0} & \RC{26.7} & \RC{0.0} & \RC{0.0} \\ 
        \text{T} & \GC{2.7} & \GC{0.0} & \GC{97.3} & \GC{0.0} & \GC{0.0} \\ 
        \text{FP} & \RC{98.7} & \RC{0.0} & \RC{1.3} & \RC{0.0} & \RC{0.0} \\ 
        \text{FT} & \RC{98.7} & \RC{0.0} & \RC{1.3} & \RC{0.0} & \RC{0.0} \end{array}$
    &
    $\begin{array}{ccccc}
        \multicolumn{1}{c}{\text{B}} & \multicolumn{1}{c}{\text{P}} & \multicolumn{1}{c}{\text{T}} & \multicolumn{1}{c}{\text{FP}} & \multicolumn{1}{c}{\text{FT}} \\ \hline
        \GC{81.3} & \GC{14.7} & \GC{4.0} & \GC{0.0} & \GC{0.0} \\ 
        \GC{16.0} & \GC{72.0} & \GC{12.0} & \GC{0.0} & \GC{0.0} \\ 
        \GC{1.3} & \GC{4.0} & \GC{94.7} & \GC{0.0} & \GC{0.0} \\ 
        \RC{90.7} & \RC{6.7} & \RC{2.7} & \RC{0.0} & \RC{0.0} \\ 
        \RC{90.7} & \RC{8.0} & \RC{1.3} & \RC{0.0} & \RC{0.0} \end{array}$
    &
    $\begin{array}{ccccc}
        \multicolumn{1}{c}{\text{B}} & \multicolumn{1}{c}{\text{P}} & \multicolumn{1}{c}{\text{T}} & \multicolumn{1}{c}{\text{FP}} & \multicolumn{1}{c}{\text{FT}} \\ \hline
        \GC{84.0} & \GC{12.0} & \GC{4.0} & \GC{0.0} & \GC{0.0} \\ 
        \GC{14.7} & \GC{73.3} & \GC{12} & \GC{0.0} & \GC{0.0} \\ 
        \GC{1.3} & \GC{4.0} & \GC{94.7} & \GC{0.0} & \GC{0.0} \\ 
        \RC{0.0} & \RC{0.0} & \RC{0.0} & \RC{0.0} & \RC{100} \\ 
        \GC{0.0} & \GC{0.0} & \GC{0.0} & \GC{0.0} & \GC{100} \end{array}$
    &
    $\begin{array}{ccccc}
        \multicolumn{1}{c}{\text{B}} & \multicolumn{1}{c}{\text{P}} & \multicolumn{1}{c}{\text{T}} & \multicolumn{1}{c}{\text{FP}} & \multicolumn{1}{c}{\text{FT}} \\ \hline
        \GC{84.0} & \GC{13.3} & \GC{1.3} & \GC{1.3} & \GC{0.0} \\ 
        \GC{24.0} & \GC{62.7} & \GC{13.3} & \GC{0.0} & \GC{0.0} \\ 
        \GC{2.7} & \GC{4.0} & \GC{93.3} & \GC{0.0} & \GC{0.0} \\ 
        \GC{0.0} & \GC{0.0} & \GC{0.0} & \GC{94.7} & \GC{5.3} \\ 
        \GC{0.0} & \GC{0.0} & \GC{0.0} & \GC{5.3} & \GC{94.7} \end{array}$
\\[1.5ex]
    M2 & M3 & M4 & M5 \\[1.5ex]

    \multicolumn{4}{c}{\textbf{TypeNet - User Independent}} \\ [1ex]
    $\begin{array}{c|ccccc}
         & \multicolumn{1}{c}{\text{B}} & \multicolumn{1}{c}{\text{P}} & \multicolumn{1}{c}{\text{T}} & \multicolumn{1}{c}{\text{FP}} & \multicolumn{1}{c}{\text{FT}} \\ \cline{2-6}
        \text{B} & \GC{98.7} & \GC{0.0} & \GC{1.3} & \GC{0.0} & \GC{0.0} \\ 
        \text{P} & \RC{72.0} & \RC{0.0} & \RC{28.0} & \RC{0.0} & \RC{0.0} \\ 
        \text{T} & \GC{5.3} & \GC{0.0} & \GC{94.7} & \GC{0.0} & \GC{0.0} \\ 
        \text{FP} & \RC{49.3} & \RC{0.0} & \RC{50.7} & \RC{0.0} & \RC{0.0} \\ 
        \text{FT} & \RC{34.7} & \RC{0.0} & \RC{65.3} & \RC{0.0} & \RC{0.0} \end{array}$
    &
    $\begin{array}{ccccc}
        \multicolumn{1}{c}{\text{B}} & \multicolumn{1}{c}{\text{P}} & \multicolumn{1}{c}{\text{T}} & \multicolumn{1}{c}{\text{FP}} & \multicolumn{1}{c}{\text{FT}} \\ \hline
        \GC{82.7} & \GC{16.0} & \GC{1.3} & \GC{0.0} & \GC{0.0} \\ 
        \GC{21.3} & \GC{66.7} & \GC{12.0} & \GC{0.0} & \GC{0.0} \\ 
        \GC{1.3} & \GC{8.0} & \GC{90.7} & \GC{0.0} & \GC{0.0} \\ 
        \RC{49.3} & \RC{12.0} & \RC{38.7} & \RC{0.0} & \RC{0.0} \\ 
        \RC{32.0} & \RC{0.0} & \RC{68.0} & \RC{0.0} & \RC{0.0} \end{array}$
    &
    $\begin{array}{ccccc}
        \multicolumn{1}{c}{\text{B}} & \multicolumn{1}{c}{\text{P}} & \multicolumn{1}{c}{\text{T}} & \multicolumn{1}{c}{\text{FP}} & \multicolumn{1}{c}{\text{FT}} \\ \hline
        \GC{72.0} & \GC{28.0} & \GC{0.0} & \GC{0.0} & \GC{0.0} \\ 
        \GC{32.0} & \GC{53.3} & \GC{14.7} & \GC{0.0} & \GC{0.0} \\ 
        \GC{1.3} & \GC{5.3} & \GC{93.3} & \GC{0.0} & \GC{0.0} \\ 
        \RC{0.0} & \RC{0.0} & \RC{0.0} & \RC{0.0} & \RC{100} \\ 
        \GC{0.0} & \GC{0.0} & \GC{0.0} & \GC{0.0} & \GC{100} \end{array}$
    &
    $\begin{array}{ccccc}
        \multicolumn{1}{c}{\text{B}} & \multicolumn{1}{c}{\text{P}} & \multicolumn{1}{c}{\text{T}} & \multicolumn{1}{c}{\text{FP}} & \multicolumn{1}{c}{\text{FT}} \\ \hline
        \GC{81.3} & \GC{17.3} & \GC{1.3} & \GC{0.0} & \GC{0.0} \\ 
        \GC{25.3} & \GC{61.3} & \GC{13.3} & \GC{0.0} & \GC{0.0} \\ 
        \GC{2.7} & \GC{4.0} & \GC{93.3} & \GC{0.0} & \GC{0.0} \\ 
        \GC{0.0} & \GC{0.0} & \GC{0.0} & \GC{82.7} & \GC{17.3} \\ 
        \GC{0.0} & \GC{0.0} & \GC{0.0} & \GC{13.3} & \GC{86.7} \end{array}$
\\[1.5ex]
    M2 & M3 & M4 & M5 \\[1.5ex]

    
    \end{tabular}
    
    } 

    \caption{Confusion Matrix for User-Independent Evaluations. Cells corresponding to explicitly trained classes utilize a white-to-blue gradient. Cells corresponding to unseen attack vectors (zero-shot evaluation) utilize a white-to-orange gradient to highlight vulnerabilities.}
    \label{fig:full_experiment_grid_UIE}
\end{figure*}

\begin{figure*}[t]
    \centering
    \renewcommand{\arraystretch}{0.83}
    \setlength{\tabcolsep}{2pt} 
    
    \resizebox{\textwidth}{!}{
    
    \begin{tabular}{c | c | c | c}
    \multicolumn{4}{c}{\textbf{Rhythmic - Context Independent}} \\ [1ex]
    $\begin{array}{c|ccccc}
         & \multicolumn{1}{c}{\text{B}} & \multicolumn{1}{c}{\text{P}} & \multicolumn{1}{c}{\text{T}} & \multicolumn{1}{c}{\text{FP}} & \multicolumn{1}{c}{\text{FT}} \\ \cline{2-6}
        \text{B} & \GC{89.8} & \GC{0.0} & \GC{10.2} & \GC{0.0} & \GC{0.0} \\
        \text{P} & \RC{72.4} & \RC{0.0} & \RC{27.6} & \RC{0.0} & \RC{0.0} \\
        \text{T} & \GC{10.2} & \GC{0.0} & \GC{89.8} & \GC{0.0} & \GC{0.0} \\
        \text{FP} & \RC{85.9} & \RC{0.0} & \RC{14.1} & \RC{0.0} & \RC{0.0} \\
        \text{FT} & \RC{36.3} & \RC{0.0} & \RC{63.7} & \RC{0.0} & \RC{0.0}
    \end{array}$
    &
    $\begin{array}{ccccc}
        \multicolumn{1}{c}{\text{B}} & \multicolumn{1}{c}{\text{P}} & \multicolumn{1}{c}{\text{T}} & \multicolumn{1}{c}{\text{FP}} & \multicolumn{1}{c}{\text{FT}} \\ \hline
        \GC{64.4} & \GC{31.1} & \GC{4.4} & \GC{0.0} & \GC{0.0} \\
        \GC{31.1} & \GC{51.6} & \GC{17.3} & \GC{0.0} & \GC{0.0} \\
        \GC{4.9} & \GC{12.9} & \GC{82.2} & \GC{0.0} & \GC{0.0} \\
        \RC{47.4} & \RC{45.9} & \RC{6.7} & \RC{0.0} & \RC{0.0} \\
        \RC{17.8} & \RC{43.0} & \RC{39.3} & \RC{0.0} & \RC{0.0}
    \end{array}$
    &
    $\begin{array}{ccccc}
        \multicolumn{1}{c}{\text{B}} & \multicolumn{1}{c}{\text{P}} & \multicolumn{1}{c}{\text{T}} & \multicolumn{1}{c}{\text{FP}} & \multicolumn{1}{c}{\text{FT}} \\ \hline
        \GC{65.8} & \GC{29.3} & \GC{4.4} & \GC{0.0} & \GC{0.4} \\
        \GC{31.6} & \GC{50.2} & \GC{18.2} & \GC{0.0} & \GC{0.0} \\
        \GC{4.4} & \GC{12.9} & \GC{82.7} & \GC{0.0} & \GC{0.0} \\
        \RC{0.0} & \RC{0.0} & \RC{0.0} & \RC{0.0} & \RC{100.0} \\
        \GC{0.0} & \GC{0.0} & \GC{0.0} & \GC{0.0} & \GC{100.0}
    \end{array}$
    &
    $\begin{array}{ccccc}
        \multicolumn{1}{c}{\text{B}} & \multicolumn{1}{c}{\text{P}} & \multicolumn{1}{c}{\text{T}} & \multicolumn{1}{c}{\text{FP}} & \multicolumn{1}{c}{\text{FT}} \\ \hline
        \GC{67.1} & \GC{29.3} & \GC{3.6} & \GC{0.0} & \GC{0.0} \\
        \GC{31.6} & \GC{52.4} & \GC{16.0} & \GC{0.0} & \GC{0.0} \\
        \GC{3.1} & \GC{13.3} & \GC{83.6} & \GC{0.0} & \GC{0.0} \\
        \GC{0.0} & \GC{0.0} & \GC{0.0} & \GC{75.6} & \GC{24.4} \\
        \GC{0.0} & \GC{0.0} & \GC{0.0} & \GC{22.2} & \GC{77.8}
    \end{array}$
\\[1.5ex]
    M2 & M3 & M4 & M5 \\[1.5ex]

\multicolumn{4}{c}{\textbf{Temporal - Context Independent}} \\ [1ex]
$\begin{array}{c|ccccc}
 & \multicolumn{1}{c}{\text{B}} & \multicolumn{1}{c}{\text{P}} & \multicolumn{1}{c}{\text{T}} & \multicolumn{1}{c}{\text{FP}} & \multicolumn{1}{c}{\text{FT}} \\ \cline{2-6}
\text{B}    & \GC{77.8} & \GC{0.0}  & \GC{22.2} & \GC{0.0} & \GC{0.0} \\
\text{P}    & \RC{55.1} & \RC{0.0}  & \RC{44.9} & \RC{0.0} & \RC{0.0} \\
\text{T}    & \GC{38.2} & \GC{0.0}  & \GC{61.8} & \GC{0.0} & \GC{0.0} \\
\text{FP} & \RC{74.2} & \RC{0.0}  & \RC{25.8} & \RC{0.0} & \RC{0.0} \\
\text{FT} & \RC{60.9} & \RC{0.0}  & \RC{39.1} & \RC{0.0} & \RC{0.0}
\end{array}$
&
$\begin{array}{ccccc}
\multicolumn{1}{c}{\text{B}} & \multicolumn{1}{c}{\text{P}} & \multicolumn{1}{c}{\text{T}} & \multicolumn{1}{c}{\text{FP}} & \multicolumn{1}{c}{\text{FT}} \\ \hline
\GC{64.4} & \GC{21.8} & \GC{13.8} & \GC{0.0} & \GC{0.0} \\
\GC{29.3} & \GC{44.0} & \GC{26.7} & \GC{0.0} & \GC{0.0} \\
\GC{18.2} & \GC{33.8} & \GC{48.0} & \GC{0.0} & \GC{0.0} \\
\RC{42.2} & \RC{49.3} & \RC{8.4}  & \RC{0.0} & \RC{0.0} \\
\RC{40.0} & \RC{40.4} & \RC{19.6} & \RC{0.0} & \RC{0.0}
\end{array}$
&
$\begin{array}{ccccc}
\multicolumn{1}{c}{\text{B}} & \multicolumn{1}{c}{\text{P}} & \multicolumn{1}{c}{\text{T}} & \multicolumn{1}{c}{\text{FP}} & \multicolumn{1}{c}{\text{FT}} \\ \hline
\GC{62.2} & \GC{22.7} & \GC{15.1} & \GC{0.0} & \GC{0.0} \\
\GC{30.7} & \GC{47.1} & \GC{22.2} & \GC{0.0} & \GC{0.0} \\
\GC{24.9} & \GC{28.0} & \GC{47.1} & \GC{0.0} & \GC{0.0} \\
\RC{0.0}  & \RC{1.3}  & \RC{0.0}  & \RC{0.0} & \RC{98.7} \\
\GC{0.0}  & \GC{0.0}  & \GC{0.0}  & \GC{0.0} & \GC{100.0}
\end{array}$
&
$\begin{array}{ccccc}
\multicolumn{1}{c}{\text{B}} & \multicolumn{1}{c}{\text{P}} & \multicolumn{1}{c}{\text{T}} & \multicolumn{1}{c}{\text{FP}} & \multicolumn{1}{c}{\text{FT}} \\ \hline
\GC{60.4} & \GC{22.2} & \GC{16.9} & \GC{0.0} & \GC{0.4} \\
\GC{29.8} & \GC{47.6} & \GC{22.7} & \GC{0.0} & \GC{0.0} \\
\GC{22.7} & \GC{28.0} & \GC{48.9} & \GC{0.4} & \GC{0.0} \\
\GC{0.0}  & \GC{0.0}  & \GC{0.0}  & \GC{66.2} & \GC{33.8} \\
\GC{0.0}  & \GC{0.0}  & \GC{0.0}  & \GC{38.2} & \GC{61.8}
\end{array}$
\\
M2 & M3 & M4 & M5 \\[1.5ex]

    \multicolumn{4}{c}{\textbf{1D-CNN - Context Independent}} \\ [1ex]
    $\begin{array}{c|ccccc}
         & \multicolumn{1}{c}{\text{B}} & \multicolumn{1}{c}{\text{P}} & \multicolumn{1}{c}{\text{T}} & \multicolumn{1}{c}{\text{FP}} & \multicolumn{1}{c}{\text{FT}} \\ \cline{2-6}
        \text{B} & \GC{99.6} & \GC{0.0} & \GC{0.4} & \GC{0.0} & \GC{0.0} \\ 
        \text{P} & \RC{77.3} & \RC{0.0} & \RC{22.7} & \RC{0.0} & \RC{0.0} \\ 
        \text{T} & \GC{3.6} & \GC{0.0} & \GC{96.4} & \GC{0.0} & \GC{0.0} \\ 
        \text{FP} & \RC{100} & \RC{0.0} & \RC{0.0} & \RC{0.0} & \RC{0.0} \\ 
        \text{FT} & \RC{97.8} & \RC{0.0} & \RC{2.2} & \RC{0.0} & \RC{0.0} \end{array}$
    &
    $\begin{array}{ccccc}
        \multicolumn{1}{c}{\text{B}} & \multicolumn{1}{c}{\text{P}} & \multicolumn{1}{c}{\text{T}} & \multicolumn{1}{c}{\text{FP}} & \multicolumn{1}{c}{\text{FT}} \\ \hline
        \GC{76.9} & \GC{21.8} & \GC{1.3} & \GC{0.0} & \GC{0.0} \\ 
        \GC{23.1} & \GC{66.7} & \GC{10.2} & \GC{0.0} & \GC{0.0} \\ 
        \GC{3.1} & \GC{10.7} & \GC{86.2} & \GC{0.0} & \GC{0.0} \\ 
        \RC{77.3} & \RC{22.7} & \RC{0.0} & \RC{0.0} & \RC{0.0} \\ 
        \RC{80.4} & \RC{19.6} & \RC{0.0} & \RC{0.0} & \RC{0.0} \end{array}$
    &
    $\begin{array}{ccccc}
        \multicolumn{1}{c}{\text{B}} & \multicolumn{1}{c}{\text{P}} & \multicolumn{1}{c}{\text{T}} & \multicolumn{1}{c}{\text{FP}} & \multicolumn{1}{c}{\text{FT}} \\ \hline
        \GC{75.6} & \GC{24.0} & \GC{0.4} & \GC{0.0} & \GC{0.0} \\ 
        \GC{25.8} & \GC{65.8} & \GC{8.4} & \GC{0.0} & \GC{0.0} \\ 
        \GC{2.2} & \GC{15.6} & \GC{82.2} & \GC{0.0} & \GC{0.0} \\ 
        \RC{0.0} & \RC{0.0} & \RC{0.0} & \RC{0.0} & \RC{100} \\ 
        \GC{0.0} & \GC{0.0} & \GC{0.0} & \GC{0.0} & \GC{100} \end{array}$
    &
    $\begin{array}{ccccc}
        \multicolumn{1}{c}{\text{B}} & \multicolumn{1}{c}{\text{P}} & \multicolumn{1}{c}{\text{T}} & \multicolumn{1}{c}{\text{FP}} & \multicolumn{1}{c}{\text{FT}} \\ \hline
        \GC{84.4} & \GC{15.1} & \GC{0.4} & \GC{0.0} & \GC{0.0} \\ 
        \GC{33.3} & \GC{54.7} & \GC{12.0} & \GC{0.0} & \GC{0.0} \\ 
        \GC{5.3} & \GC{10.2} & \GC{84.4} & \GC{0.0} & \GC{0.0} \\ 
        \GC{0.0} & \GC{0.0} & \GC{0.0} & \GC{89.8} & \GC{10.2} \\ 
        \GC{0.0} & \GC{0.0} & \GC{0.0} & \GC{15.6} & \GC{84.4} \end{array}$
    \\
    M2 & M3 & M4 & M5 \\[1.5ex]

    \multicolumn{4}{c}{\textbf{TypeNet - Context Independent}} \\ [1ex]
    $\begin{array}{c|ccccc}
         & \multicolumn{1}{c}{\text{B}} & \multicolumn{1}{c}{\text{P}} & \multicolumn{1}{c}{\text{T}} & \multicolumn{1}{c}{\text{FP}} & \multicolumn{1}{c}{\text{FT}} \\ \cline{2-6}
        \text{B} & \GC{98.7} & \GC{0.0} & \GC{1.3} & \GC{0.0} & \GC{0.0} \\ 
        \text{P} & \RC{77.3} & \RC{0.0} & \RC{22.7} & \RC{0.0} & \RC{0.0} \\ 
        \text{T} & \GC{3.6} & \GC{0.0} & \GC{96.4} & \GC{0.0} & \GC{0.0} \\ 
        \text{FP} & \RC{82.7} & \RC{0.0} & \RC{17.3} & \RC{0.0} & \RC{0.0} \\ 
        \text{FT} & \RC{60.0} & \RC{0.0} & \RC{40.0} & \RC{0.0} & \RC{0.0} \end{array}$
    &
    $\begin{array}{ccccc}
        \multicolumn{1}{c}{\text{B}} & \multicolumn{1}{c}{\text{P}} & \multicolumn{1}{c}{\text{T}} & \multicolumn{1}{c}{\text{FP}} & \multicolumn{1}{c}{\text{FT}} \\ \hline
        \GC{72.4} & \GC{27.1} & \GC{0.4} & \GC{0.0} & \GC{0.0} \\ 
        \GC{20.4} & \GC{68.9} & \GC{10.7} & \GC{0.0} & \GC{0.0} \\ 
        \GC{3.1} & \GC{16.9} & \GC{80.0} & \GC{0.0} & \GC{0.0} \\ 
        \RC{67.1} & \RC{26.7} & \RC{6.2} & \RC{0.0} & \RC{0.0} \\ 
        \RC{60.9} & \RC{22.2} & \RC{16.9} & \RC{0.0} & \RC{0.0} \end{array}$
    &
    $\begin{array}{ccccc}
        \multicolumn{1}{c}{\text{B}} & \multicolumn{1}{c}{\text{P}} & \multicolumn{1}{c}{\text{T}} & \multicolumn{1}{c}{\text{FP}} & \multicolumn{1}{c}{\text{FT}} \\ \hline
        \GC{76.0} & \GC{23.6} & \GC{0.4} & \GC{0.0} & \GC{0.0} \\ 
        \GC{24.4} & \GC{64.0} & \GC{11.6} & \GC{0.0} & \GC{0.0} \\ 
        \GC{2.7} & \GC{13.3} & \GC{84.0} & \GC{0.0} & \GC{0.0} \\ 
        \RC{0.0} & \RC{0.0} & \RC{0.0} & \RC{0.0} & \RC{100} \\ 
        \GC{0.0} & \GC{0.0} & \GC{0.0} & \GC{0.0} & \GC{100} \end{array}$
    &
    $\begin{array}{ccccc}
        \multicolumn{1}{c}{\text{B}} & \multicolumn{1}{c}{\text{P}} & \multicolumn{1}{c}{\text{T}} & \multicolumn{1}{c}{\text{FP}} & \multicolumn{1}{c}{\text{FT}} \\ \hline
        \GC{68.0} & \GC{30.2} & \GC{1.8} & \GC{0.0} & \GC{0.0} \\ 
        \GC{23.6} & \GC{67.1} & \GC{9.3} & \GC{0.0} & \GC{0.0} \\ 
        \GC{1.3} & \GC{14.2} & \GC{84.4} & \GC{0.0} & \GC{0.0} \\ 
        \GC{0.0} & \GC{0.0} & \GC{0.0} & \GC{93.3} & \GC{6.7} \\ 
        \GC{0.0} & \GC{0.0} & \GC{0.0} & \GC{31.6} & \GC{68.4} \end{array}$
    \\
    M2 & M3 & M4 & M5
    
    \end{tabular}
    
    } 

    \caption{Confusion matrices for context-independent evaluation. Cells corresponding to explicitly trained classes utilize a white-to-blue gradient. Cells corresponding to unseen attack vectors (zero-shot evaluation) utilize a white-to-orange gradient to highlight vulnerabilities.}
    \label{fig:full_experiment_grid_CIE}
\end{figure*}
\subsubsection{Training and evaluation}

Models are trained and evaluated under two complementary generalization settings, user-independent evaluation (UIE) and context-independent evaluation (CIE), $\pi \in \{\text{UIE}, \text{CIE}\}$, capturing variability across users and prompts.

In the \textit{user-independent evaluation} (UIE), training and testing are performed on disjoint sets of users. This setting evaluates whether $f_\theta$ generalizes to unseen individuals, reflecting deployment scenarios where no prior behavioral data is available. In the \textit{context-independent evaluation} (CIE), training and testing are performed on disjoint sets of prompts, with overlap allowed in users. This setting isolates generalization across tasks, ensuring that predictions rely on stable behavioral signals rather than prompt-specific artifacts.

Within each setting, four training configurations are defined to examine the effect of exposure to assisted writing and adversarial behavior. Let $\mathcal{Y} = \{B, T, P, FT, FP\}$. For each configuration $m \in \{M2, M3, M4, M5\}$, training is restricted to a subset $\mathcal{Y}^{m}_{train} \subseteq \mathcal{Y}$, while evaluation is performed over all classes.

$M2$ uses $\{B, T\}$ and tests whether exposure to transcription alone is sufficient to detect more complex behaviors.  
$M3$ uses $\{B, T, P\}$ and evaluates whether modeling paraphrasing improves robustness to LLM-assisted writing.  
$M4$ uses $\{B, T, P, FT\}$ and assesses zero-shot adversarial generalization by exposing the model to one form of behavioral manipulation.  
$M5$ uses all classes and represents the fully supervised setting in which all behaviors are observed during training.

To obtain robust estimates, 3-fold cross-validation is applied in both settings. In UIE, users are partitioned into three disjoint groups, with two used for training and one for testing. In CIE, prompts are partitioned into three groups with balanced cognitive load, and models are trained on two groups and evaluated on unseen prompts. This design preserves variability while preventing overlap between training and testing distributions.

Hyperparameters are optimized using nested cross-validation within each training fold to avoid data leakage. For each outer split, the training data is further divided into inner training and validation subsets, and model parameters are fine-tuned.

Performance is evaluated using macro-averaged F1 (Macro-F1), False Acceptance Rate (FAR), and False Rejection Rate (FRR). Macro-F1 is the unweighted mean of the per-class F1 scores, where F1 is the harmonic mean of precision and recall, computed over all five classes; it therefore weights each writing mode equally regardless of the number of samples. For FAR and FRR, bona fide samples ($B$) are treated as the positive class, while all assisted or adversarial samples are treated as negative. FAR measures the proportion of assisted responses incorrectly classified as bona fide, while FRR measures the proportion of bona fide responses incorrectly rejected. All metrics are computed from confusion matrices pooled over the three folds.

\section{Results and Discussion}
\label{sec:results_discussion}

We evaluate the effectiveness of keystroke dynamics in distinguishing bona fide writing from multiple forms of LLM-assisted behavior under both standard and adversarial conditions. Across all experimental settings, keystroke signals provide a strong and consistent basis for discrimination, particularly when modeled using sequential architectures. Table~\ref{tab:m5_summary} summarizes the performance of all four models in the M5 configuration.

\begin{table}[t]
\caption{M5 results (\%), pooled over three folds. Macro-F1 averages the per-class F1; FAR/FRR treat bona fide ($B$) as positive. Best per row in bold.}
\label{tab:m5_summary}
\centering
\renewcommand{\arraystretch}{1.1}
\begin{tabular}{llcccc}
\toprule
 & Metric & Temporal & Rhythmic & 1D-CNN & TypeNet \\
\midrule
\multirow{3}{*}{UIE} & Macro-F1 $\uparrow$ & 58.8 & 67.7 & \textbf{85.6} & 80.8 \\
 & FAR $\downarrow$ & 13.3 & 12.8 & \textbf{6.7} & 7.0 \\
 & FRR $\downarrow$ & 45.3 & 40.0 & \textbf{16.0} & 18.7 \\
\midrule
\multirow{3}{*}{CIE} & Macro-F1 $\uparrow$ & 56.9 & 71.2 & \textbf{79.3} & 76.3 \\
 & FAR $\downarrow$ & 13.1 & 10.8 & 9.7 & \textbf{6.2} \\
 & FRR $\downarrow$ & 39.6 & 32.9 & \textbf{15.6} & 32.0 \\
\bottomrule
\end{tabular}
\end{table}

\vspace{0.5em}
\noindent\underline{\textbf{Model comparison and overall performance}} Sequential models, especially the 1D-CNN, consistently outperform feature-based approaches across both user-independent and context-independent evaluations. The 1D-CNN achieves a false acceptance rate (FAR) of 6.7\% in the user-independent setting, while maintaining 84.0\% classification accuracy for bona fide samples in the fully supervised M5 setting. In contrast, temporal feature-based representations achieve 54.7\% and 60.4\% bona fide accuracy under UIE and CIE, respectively, compared with 84.0\% and 84.4\% for the 1D-CNN. These differences highlight a key finding: discriminative information in keystroke dynamics is not adequately captured by summary statistics alone but is distributed across temporal structure. Sequential models can exploit this structure, particularly local dependencies such as pause-burst patterns and revision behavior.

\vspace{0.5em}
\noindent\underline{\textbf{Separability across writing modes}} A more detailed analysis shows clear and consistent separability between transcription and bona fide writing. Under the 1D-CNN, transcription is classified with accuracy above 93\% in the user-independent setting, with misclassification into the bona fide class typically below 5\%. This demonstrates that transcription is behaviorally distinct in the learned keystroke representations, consistent with reduced variability and limited revision activity. In contrast, paraphrasing presents a more challenging case. Under the 1D-CNN, when paraphrasing is not observed during training (M2), 73.3\% of paraphrased samples are classified as bona fide in the user-independent setting. When paraphrasing is included during training (M3), classification accuracy increases to 72.0\%, although 16.0\% of paraphrased samples are still misclassified as bona fide. Transcription remains substantially easier to distinguish, with 94.7\% classification accuracy under the same M3 configuration.

This behavior reflects a fundamental property of the task. Paraphrasing shares key behavioral signals associated with genuine writing, including timing variability and revision patterns, thereby reducing separability and increasing overlap with bona fide distributions.

\vspace{0.5em}
\noindent\underline{\textbf{Adversarial behavior and robustness}} Behaviorally manipulated samples derived from paraphrasing and transcription further stress the detection framework. In zero-shot settings without adversarial exposure, these samples are frequently misclassified as bona fide. For example, under the 1D-CNN in the user-independent M2 configuration, approximately 98.7\% of adversarial samples are assigned to the bona fide class, indicating that such manipulations can closely align assisted writing with the statistical characteristics of genuine composition.

At the same time, these results highlight an important strength of the framework. When adversarial samples are incorporated during training, performance improves substantially. In the M5 setting of 1D-CNN UIE, adversarial classes are classified with accuracy above 85\%, corresponding to improvements exceeding 80 percentage points relative to the zero-shot case. This demonstrates that adversarial modeling provides an effective defense: exposure to behaviorally grounded variations enables the model to recover separability and maintain robust performance under realistic threat conditions. Importantly, this improvement reflects adaptation to observed behavioral patterns, emphasizing that robustness depends on coverage of realistic writing behaviors rather than static model capacity.

\vspace{0.15em}
\noindent\underline{\textbf{Role of feature representations}} Temporal features degrade significantly under cognitive variation, with reduced accuracy and increased confusion between bona fide and assisted writing. Rhythmic features provide partial improvement by capturing higher-level structure such as pauses and bursts, but still exhibit overlap between paraphrasing and bona fide writing. Sequential models maintain more stable performance across most classes, indicating that effective detection requires modeling keystroke behavior across multiple temporal scales. This supports the use of sequence-based representations as a primary modeling approach for this task.

\vspace{0.5em}
\noindent\underline{\textbf{Generalization across users and contexts}} These trends remain consistent across both evaluation settings. In the user-independent setting, models perform well on unseen individuals, indicating that the learned representations capture task-dependent behavior rather than user-specific traits. In the context-independent setting, performance remains relatively strong on unseen prompts, particularly for sequential models, suggesting that the learned signal is not entirely prompt-specific.

\vspace{0.5em}
\noindent\underline{\textbf{Error structure and behavioral interpretation}} The confusion matrices reveal a consistent and interpretable error structure aligned with behavioral similarity. Transcription remains well separated from bona fide writing, while paraphrasing and adversarial samples cluster closer to the bona fide class when not explicitly modeled. This pattern reflects differences in cognitive demand: transcription involves lower cognitive effort and remains distinct, whereas paraphrasing and behavioral manipulation increase variability and move closer to genuine writing behavior.

Overall, the results show that keystroke-based detection captures differences in writing behavior at the level of cognitive effort and interaction. While paraphrasing and adversarial manipulation reduce separability by mimicking behavioral signals, the framework remains effective when trained on sufficiently diverse data.

\vspace{0.5em}
\noindent\underline{\textbf{Ethical considerations}} Keystroke dynamics constitutes behavioral data and should therefore be collected with informed consent, appropriate privacy safeguards, and limited data retention. Predictions from the proposed framework should not be treated as definitive evidence of academic misconduct, but rather as an additional signal that may support further review.

\vspace{0.5em}
\noindent\underline{\textbf{Limitations}} 
The study has several limitations. The dataset includes $45$ participants, with $30$ returning for the second phase, and is limited to Vietnamese writing. In addition, the adversarial samples are behaviorally motivated transformations rather than data collected from users actively attempting to evade detection. Future work will evaluate the framework with larger and more diverse populations, additional languages, and human-generated adversarial behaviors. 
\section{Conclusion and Future Work}
\label{sec:conclusion_and_futurework}

We investigated keystroke dynamics as a behavioral signal to detect LLM-assisted writing under both realistic and adversarial conditions. Our results show that keystroke signals capture meaningful differences in the writing process, particularly for transcription. Sequential models provide consistent gains over feature-based approaches across both user-independent and context-independent settings. At the same time, the findings expose clear limitations. Detection is not uniform across writing behaviors: paraphrasing remains difficult to distinguish from bona fide writing, and behaviorally manipulated samples were misclassified when not represented in the training set. These patterns indicate that performance is driven less by inherent separability and more by coverage of writing behaviors during training.

We further demonstrated that training with behaviorally grounded adversarial samples substantially improves robustness to the modeled adversarial behaviors, enabling models to recover strong separability for adversarial samples that are frequently misclassified as bona fide in zero-shot settings. This suggests that keystroke-based detection should be viewed as an evolving system rather than a static classifier. In practice, we envision this framework as a complementary second layer following conventional similarity-based plagiarism detection systems rather than as a replacement. Behavioral analysis is particularly valuable for submissions that exhibit little textual overlap despite possible LLM assistance.

In the future, we plan to scale the dataset to larger, more diverse populations to validate its stability across age groups, devices, and typing proficiency levels. We also plan to study more realistic adversarial settings, including adaptive users who iteratively respond to detector feedback, to better approximate deployment conditions. Finally, we plan to investigate complementary behavioral signals, such as the acoustic characteristics of keystrokes and cursor movements, to determine whether they provide additional information for detecting LLM-assisted writing.
\balance
\bibliography{references}

\end{document}